\documentclass[graybox, envcountchap]{svmult}

\usepackage{import}

\usepackage{makeidx}         
\usepackage{graphicx}        
\usepackage{multicol}        
\usepackage[bottom]{footmisc}

\usepackage{wrapfig}
\usepackage{newtxtext}       %
\usepackage{newtxmath}       

\usepackage{color}
\usepackage{dsfont}
\usepackage{comment}
\usepackage{url}
\usepackage{lscape}
\usepackage{arydshln}
\usepackage{todonotes}
\usepackage{algorithmicx}
\usepackage[linesnumbered,ruled,vlined]{algorithm2e}
\usepackage{algpseudocode}
\usepackage[square,numbers,sectionbib]{natbib}
\usepackage{booktabs}
\usepackage{comment}
\usepackage{bm}

\definecolor{mygreen}{RGB}{12, 141, 85}
\definecolor{mypurple}{RGB}{102, 0, 204}

\renewcommand{\E}{\operatorname{E}}

\newcommand{\BF}[1]{
	\relax
	\ifmmode
	\ifcat\noexpand#1\relax 
		\boldsymbol{#1}     
	\else
		\mathbf{#1}
	\fi
	\else
		\textbf{#1}
	\fi
}

\makeindex             

\usepackage{array}
\newcolumntype{P}[1]{>{\centering\arraybackslash}m{#1}}
\usepackage{todonotes}

\usepackage{type1cm}        
\usepackage{makeidx}         
\usepackage{graphicx}        
\usepackage{multicol}        
\usepackage[bottom]{footmisc}

\usepackage{amsmath}
\usepackage{amsfonts}
\usepackage{enumerate}

\usepackage{newtxtext}       %
\usepackage{newtxmath}       

\usepackage{comment}

\usepackage{glossaries}

\usepackage{glossaries}

\makeglossaries

\newglossaryentry{gloss:heterogeneous objectives}
{
    name=heterogeneous objectives,
    description={Objective function components of a problem that differ in complexity, latency, uncertainty, domain type, co-domain type, black box vs analytically known, or in other ways that might affect search}
}
\newglossaryentry{gloss:Waiting strategy}
{
    name=Waiting strategy,
    description={A technique for handling heterogeneous objectives differing in latency equating to simply waiting for the slower objective(s) to be evaluated while the faster objective(s) have already been evaluate}
}
\newglossaryentry{gloss:uncertain objective function}
{
    name=uncertain objective function,
    description={An objective function giving a stochastic output for a given input (decision vector)},
}
\newglossaryentry{gloss:latency}
{
    name= latency,
   description={The time taken for a process, in particular a function evaluation, see \gls{gloss:heterogeneous objectives}},
}
\newglossaryentry{gloss:asynchronous evolutionary algorithms}
{
    name= asynchronous evolutionary algorithms,
   description={Evolutionary algorithms that work on parallel architectures supporting asynchronous solution evaluations, selection and search operators},
}
\newglossaryentry{gloss:closed-loop optimization}
{
    name= closed-loop optimization,
   description={Optimization by directly connecting (e.g. physical, chemical or biological) experimental rigs to computer-based search},
}
\newglossaryentry{gloss:trust region method}
{
    name= trust region method,
   description={An optimization technique based on approximating cost function regions, usually with a quadratic},
}
\newglossaryentry{gloss:Brood Interleaving strategy}
{
    name=Brood Interleaving strategy,
   description={A technique for handling \gls{gloss:heterogeneous objectives} differing in latency that performs evaluations on slow and fast objectives in parallel, but more on the fast objective, and interleaves the information from both to assign fitness},
}
\newglossaryentry{gloss:Speculative Interleaving strategy}
{
    name=Speculative Interleaving strategy,
   description={A form of \gls{gloss:Interleaving strategy} based on looking ahead at how a solution that has not been evaluated fully yet might perform},
}
\newglossaryentry{gloss:Interleaving strategy}
{
    name=Interleaving strategy,
   description={A technique for handling \gls{gloss:heterogeneous objectives} differing in latency that performs evaluations on slow and fast objectives in parallel, but more on the fast objective, and interleaves the information from both to assign fitness},
}
\newglossaryentry{gloss:Fast-First strategy}
{
    name=Fast-First strategy,
   description={A technique for handling \gls{gloss:heterogeneous objectives} differing in latency equating to basing fitness assignment on only the fast objective(s) of a problem for most of the optimization run, using the slow objective only at the end of the run},
}
\newglossaryentry{gloss:ephemeral resource constraint}
{
    name=ephemeral resource constraint,
    description={In \gls{gloss:closed-loop optimization}, a pseudo-constraint that exists only at runtime that prevents a solution from being evaluated because a certain resource needed for the evaluation is depleted (temporarily)},
}
\newglossaryentry{gloss:interruption}
{
    name=interruption,
    description={In \gls{gloss:closed-loop optimization}, an interruption to the usual progress of the evolutionary algorithm at runtime due to an active \gls{gloss:ephemeral resource constraint}. Such interruptions prevent solutions from being evaluated and hence introduce a bias to the search that needs to be accounted for},
}
\newglossaryentry{gloss:safe optimization}
{
    name=safe optimization,
    description={Safe optimization refers to the use of strategies to mitigate losses (of life, resources) when optimizing configurable entities that can be damaged by some combinations of parameter settings (or some allele combinations)},
}
\newglossaryentry{gloss:surrogate-assisted evolutionary algorithm}
{
    name=surrogate-assisted evolutionary algorithm,
    description={A method using fitness approximation techniques in place of some objective function evaluations, typically based on neural networks, Gaussian processes, radial basis functions, or similar},
}

\usepackage{colortbl}
\usepackage{enumerate}

\definecolor{aoenglish}{rgb}{0.0, 0.5, 0.0}

\usepackage{hyperref}

\usepackage{pgfplots}
\usepackage{tikz}
\usetikzlibrary{calc, intersections}

\newcommand{\rev}[1]{}

\renewcommand{\todo}[1]{}
\newcommand{\changed}[1]{#1}
\usepackage{siunitx}

\usepackage{caption}
\usepackage{subcaption}
\usepackage{nicefrac}

\newcommand{\hide}[1]{}

\usepackage{xcolor}

\definecolor{todocolor}{rgb}{0.9,0.1,0.1}
\definecolor{changedcolor}{rgb}{0.42,0.27,0.57}

\newcommand{\nbc}[3]{
		{\colorbox{#3}{\bfseries\sffamily\scriptsize\textcolor{white}{#1}}}
		{\textcolor{#3}{\sf\small$\blacktriangleright$\textit{#2}$\blacktriangleleft$}}
}

\renewcommand{\todo}[1]{\nbc{TODO}{#1}{todocolor}}
\renewcommand{\changed}[1]{\nbc{CHANGED}{#1}{changedcolor}}

\renewcommand{\todo}[1]{}
\renewcommand{\changed}[1]{#1}

\begin{document}

\frontmatter

\mainmatter

%
%
%


\title*{Identifying Properties of Real-World Optimisation Problems through a Questionnaire}

\titlerunning{Identifying Properties of Real-World Optimisation Problems through a Questionnaire}
\author{Koen van der Blom,
    Timo M. Deist,
    Vanessa Volz,
    Mariapia Marchi,
    Yusuke Nojima,
    Boris Naujoks,
    Akira Oyama and
    Tea Tu\v{s}ar
}
\authorrunning{K. van der Blom et al.}
%
\institute{Koen van der Blom \at Leiden University, Leiden, The Netherlands, \email{k.van.der.blom@liacs.leidenuniv.nl}
\and Timo M. Deist \at Centrum Wiskunde \& Informatica, Amsterdam, The Nederlands, \email{timo.deist@cwi.nl}
\and Vanessa Volz \at modl.ai, Copenhagen, Denmark, \email{vanessa@modl.ai}
\and Mariapia Marchi \at ESTECO SpA, Trieste, Italy, \email{marchi@esteco.com}
\and Yusuke Nojima \at Osaka Prefecture University, Sakai, Osaka, Japan, \email{nojima@cs.osakafu-u.ac.jp}
\and Boris Naujoks \at TH K\"{o}ln, Gummersbach, Germany, \email{boris.naujoks@th-koeln.de}
\and Akira Oyama \at Japan Aerospace Exploration Agency, Sagamihara, Japan, \email{oyama@flab.isas.jaxa.jp}
\and Tea Tu\v{s}ar \at Jo\v{z}ef Stefan Institute, Ljubljana, Slovenia, \email{tea.tusar@ijs.si}
}
%
%
\maketitle


\abstract*{Optimisation algorithms are commonly compared on benchmarks to get insight into performance differences. However, it is not clear how closely benchmarks match the properties of real-world problems because these properties are largely unknown. This work investigates the properties of real-world problems through a questionnaire to enable the design of future benchmark problems that more closely resemble those found in the real world.
The results, while not representative \todo{R1-1}\changed{as they are based on only 45 responses, indicate} that many problems possess at least one of the following properties: they are constrained, deterministic, have only continuous variables, require substantial computation times for both the objectives and the constraints, or allow a limited number of evaluations.
Properties like known optimal solutions and analytical gradients are rarely available, limiting the options in guiding the optimisation process. These are all important aspects to consider when designing realistic benchmark problems. 
\changed{At the same time, the design of realistic benchmarks is difficult, because objective functions are often reported to be black-box and many problem properties are unknown.}
To further improve the understanding of real-world problems, readers working on a real-world optimisation problem are encouraged to fill out the questionnaire: \url{https://tinyurl.com/opt-survey}
}

\abstract{Optimisation algorithms are commonly compared on benchmarks to get insight into performance differences. However, it is not clear how closely benchmarks match the properties of real-world problems because these properties are largely unknown. This work investigates the properties of real-world problems through a questionnaire to enable the design of future benchmark problems that more closely resemble those found in the real world.
The results, while not representative \todo{R1-1}\changed{as they are based on only 45 responses, indicate} that many problems possess at least one of the following properties: they are constrained, deterministic, have only continuous variables, require substantial computation times for both the objectives and the constraints, or allow a limited number of evaluations.
Properties like known optimal solutions and analytical gradients are rarely available, limiting the options in guiding the optimisation process. These are all important aspects to consider when designing realistic benchmark problems. 
\changed{At the same time, the design of realistic benchmarks is difficult, because objective functions are often reported to be black-box and many problem properties are unknown.}
To further improve the understanding of real-world problems, readers working on a real-world optimisation problem are encouraged to fill out the questionnaire: \url{https://tinyurl.com/opt-survey}
}


\subsection*{Key words (3 to 5)}

Real-world optimisation problems, Problem properties, Questionnaire, Benchmarking


\section{Introduction}
\label{sec:intro}

Optimisation algorithms are ultimately used to solve real-world problems. In contrast, the benchmarks used in academic research to assess algorithmic performance often consist of artificial mathematical functions. As previous studies have indicated \cite{WatBarHowWhi1999}, it is questionable whether the performance of algorithms on such benchmarks translates to real-world problems.

While artificial mathematical functions are designed to have specific properties, it is not clear how closely they reflect the properties seen in the real world. In fact, work by Ishibuchi et al.~\cite{IshHeSha2019cec} points out that differences in algorithm performance exist between commonly used benchmark functions and real-world problems. In addition, Tanabe et al.~\cite{TanOya2017cec} establish that both C-DTLZ problems and common test problems that aim to imitate real-world problems contain a number of unnatural characteristics. Further, it has been shown that in some cases, simple algorithms can be surprisingly effective on real-world-like problems, in contrast to what artificial benchmarks would predict \cite{WatBarHowWhi1999}.

As a consequence, \changed{the design of benchmarks would benefit from a better understanding of which properties appear in real-world problems, how common these properties are, and in which combinations they are found.}
Further, it is of particular interest to identify characteristics of real-world problems that are not yet represented in artificial benchmarks, and to identify real-world problems that might be usable as part of a benchmark suite. All these aspects would then provide better guidance to the development of algorithms that perform well in the real world.

Although such reality-inspired benchmarks may still, in part, consist of artificial functions, those functions are then designed based on clear indications of needs. In addition, actual real-world problems or simplified problems that are correlated with them could be included in benchmarks.
By contributing problems to benchmarks, \todo{grammar}\changed{industry} can benefit from better algorithms and solutions. At the same time, academia benefits from greater insight into the particularities of the specific problem.
While understandably not all real-world problems are publicly available, some properties might still be made available, and could thus guide the design of useful benchmark problems. 

A better understanding of real-world problems and their properties is clearly important. This is why we have designed a questionnaire to gather information on real-world problems faced by researchers and practitioners. \todo{R2-1}\changed{The questionnaire was designed by researchers from the Evolutionary Computation community, which may have induced biases in questionnaire design and the reached audience. We hope that the results obtained are either directly or indirectly useful for other communities as well, even though they might have added, changed, or formulated some questions differently. Most importantly, each reader is strongly encouraged to fill out the questionnaire themselves and to advertise it in their own community.}

This chapter presents the questionnaire, and analyses and discusses the responses we have collected between October 2019 and July 2020. \changed{Our motivation for creating the questionnaire was to (a) identify properties of existing real-world problems and (b) identify or create practical benchmarks that resemble these properties.}

Based on our \changed{findings}, we are able to make some first (tentative) suggestions for constructing future benchmarking suites. For example, we find that many real-world problems include constraints and there are several with high-dimensional search spaces. 
\todo{grammar}\changed{Currently}, such problems are not well reflected in popular benchmark \todo{E-9}\changed{suites}. However, to make benchmark \todo{E-9}\changed{suites} better suited to match real-world problems, these should be adapted to reflect the above observations. What could be used directly in benchmark suites are problems with short evaluation times ($< 1$ second). We identified several of such problems.

Since benchmarks generally measure the performance of algorithms on clearly defined problems, this work is primarily focusing on the analysis of real-world problems that have already been encoded into decision variables, and corresponding objective and constraint functions. While this work does not directly make it easier for practitioners to decide which algorithm to use, it aims to do so indirectly by providing practical means for the characterisation of optimisation problems that can be used to identify suitable benchmarks. Naturally, when characteristics found through the questionnaire are not present in existing benchmarks, these benchmarks \todo{R1-1}\changed{can} be extended to accommodate practitioners with such problems.

Having clarified the \todo{R1-1}\changed{intended} relevance of this work to benchmarking, this chapter is also valuable for the general development of optimisation algorithms. Analysing responses to the questionnaire can reveal that certain research directions receive less attention than would be warranted\todo{grammar}\changed{,} given that they address frequently occurring issues in real-world problems.
For example, we find that work specialising in handling expensive optimisation problems (such as surrogate-assisted optimisation \cite{Jin2011}) is highly relevant, as several responses to the questionnaire report objective/constraint evaluation times of more than one day.
As such, the questionnaire can both inspire new research avenues and revitalise existing research directions. In turn this should result in more algorithms with properties specifically useful for real-world applications.

This chapter starts with an overview of related work in Section~\ref{sec:related}, with a focus on real-world problems in existing benchmarks. The questionnaire, designed to learn more about the properties of real-world problems, is introduced in Section~\ref{sec:questionnaire}.
Next, Section~\ref{sec:results} presents and analyses the currently available answers to the questionnaire. To conclude, the work is briefly summarised and the main points for future investigation are highlighted in Section~\ref{sec:conclusion}.

\section{Related work}
\label{sec:related}

Shortcomings of existing test suites in \todo{R2-10}\changed{Evolutionary Computation} have already been remarked more than 20 years ago \cite{WhiRanDzuMat1995}, and new suites have been proposed since then. However, even though it also has been demonstrated that algorithms behave differently on artificial and real-world(-like) problems \cite{IshHeSha2019cec,TanOya2017cec,WatBarHowWhi1999}, surprisingly little work has been done to address this issue.

Some years back, a questionnaire about benchmarks was issued in the genetic programming (GP) community \cite{WhiMcDCasMan2013}. 
One of the main results was the construction of a \todo{R2-6}\changed{list} of common benchmark problems with critical flaws, and the proposal of replacement problems (without these flaws). The study focused on challenging problems and diversity in evaluation time. Although it was useful to see how other questionnaires in the optimisation field were conducted, the aim was markedly different. 
Their work surveyed the GP community for opinions on their benchmarking practices, whereas this paper aims to ask optimisation experts from both industry and academia for the properties of their real-world problems.

In 2013, 17 engineers at UK companies were interviewed about the use of computational optimisation~\cite{TiwNorHutTur2015}. While that questionnaire focused on detecting current industrial optimisation practices and needs, some topics are shared by our questionnaire and its goals, i.e. features of industrial optimisation problems and highlighting existing issues as well as areas for further investigation.

Recently, several benchmark suites have been proposed that aim to resemble some aspects of real-world problems. A collection of design problems in engineering is presented in \cite{TanIsh2020asc}
as a problem suite for multi-objective optimisation algorithms. These problems have all been presented in the literature before and were collected into an easy-to-use framework in order to facilitate comparisons. To limit the computational resources required for evaluating each solution, the problems in this function suite are all evaluated by combinations of human-understandable functions designed by domain experts. This results in problems that are both reasonably quick to evaluate and more representative of real-world problems than conventional artificial test problems. However, by avoiding simulations and using simplified models instead, these problems cannot accurately represent some important characteristics (e.g. noise) of real-world problems.

In contrast, the recently proposed benchmarking suites mentioned in the following use simulation-based evaluations, thus accepting larger computational costs. The CFD problem suite proposed in \cite{DanRahEveTab2018ppsn} contains several optimisation problems inspired by applications in engineering that require simulations with computational fluid dynamics (CFD). 
It consists of three problems (two single-objective, one bi-objective). The GBEA
(Game-Benchmark for Evolutionary Algorithms) \cite{VolNauKerTus2019gecco} contains two single- and two multi-objective problem suites, where most of the included problems also require simulations to evaluate a given solution. In this case, the functions contained in the benchmark are optimisation problems that arise in game design.

\changed{Given} further insights, it could be determined how much of the space of real-world problems is actually covered by the benchmarks described above. For example, they do not represent the full complexity and computational costs of some optimisation problems observed in the real world due to practical constraints. Evaluations of the benchmark functions take up to 20 minutes (CFD suite) and a few seconds (GBEA). While evaluation time in benchmarks should ideally be kept short and is not necessarily an indicator of problem complexity, this difference does show that there is need for further investigation.

Besides the benchmarks listed above, there have also been several competitions featuring real-world problems. Some feature simulation-based evaluations, such as the \emph{Competition on Online Data-Driven Multi-Objective Optimization} at CEC 2019~\cite{WanHeTiaJin2019}.
The Japanese Society for Evolutionary Computation holds competitions based on real-world design optimisation problems in conjunction with the Evolutionary Computation Symposium since 2017. 
These benchmark problems include the car structure design optimisation problem provided by the Mazda motor corporation in 2017  \cite{JSEJAX2017,KohKemAkiTat2018gecco}, the Moon landing site selection problem provided by the Japan Aerospace Exploration Agency in 2018 \cite{JSEJAX2018,OyaFukTat2019}, and the wind turbine design problem provided by Hitachi, Ltd. in 2019 \cite{JSEC2019}.
Others rely on more abstract computational models of the real-world, such as the \emph{CEC 2011 Competition on Testing Evolutionary Algorithms on Real-World Optimization Problems} \cite{DasSug2010}. While we commend the efforts to include real-world problems in competitions, it is difficult to interpret the results, answer open research questions and draw conclusions for future research. Even if the problems are disclosed, their characteristics remain unclear. In addition, results are usually only published for a specific setting, e.g. for one specific budget and one specific search space dimension.

Some researchers have also made their real-world problems available~\cite{BroHorKnoLos2016}.
Examples include \todo{R2-7}\changed{radar waveform optimisation \cite{Hug2007emo}, airfoil optimisation \cite{BeuNauEmm2007ejor}, and car structure design optimisation \cite{KohKemAkiTat2018gecco}}. \todo{RP}\changed{While this represents a beginning, a few self-contained problems offer little to improve the} understanding of algorithm behaviour. These problems should be combined into a common framework to reduce the obstacles to using them, and making the results more readily comparable.

Besides the real-world problems mentioned above, where most are design problems encountered in industry, meta-problems have also been suggested as benchmark sources. Suggestions include hyperparameter optimisation \cite{DoeDreKer2019gecco} and optimisation problems from machine learning, such as tuning the weights of neural networks~\cite{EggFeuKleFal2016,KerGalPreTey2019}.
It is, however, questionable whether these insights can be translated to more conventional real-world problems, e.g. in engineering.

\section{Questionnaire}
\label{sec:questionnaire}

This section provides an overview of the questionnaire design and briefly introduces its structure.

\subsection{Background}
The idea for the questionnaire
originates from the Lorentz Center MACODA (Many Criteria Optimization and Decision Analysis) workshop that was held in Leiden, the Netherlands, in September 2019. 

As the questionnaire is intended to gather examples of real-world optimisation problems rather than a representative sample, and because we are asking for facts instead of opinions, many of the potential pitfalls in questionnaire design can be avoided in this instance. Still, we followed guidelines from \cite{KitPfl2002}, went through several iterations of the questionnaire and followed up with participants. We aimed to keep the questionnaire practical by maintaining it at a manageable length. An effort was also made to use well-known terminology and visualisations to keep things understandable for our target audience, \todo{RP}\changed{who} are optimisation experts from industry and academia.

The questionnaire was implemented online using Google Forms.
Its first version was made available to the public on \todo{RP}\changed{16 October} 2019. It was advertised through presentations at events attended by the authors and personal emails to researchers and practitioners in the field.
The preliminary results based on 21 answers to this initial questionnaire are presented in a short paper~\cite{BloDeiTusMar2020gecco}. 
This relatively small number of answers is not representative as most responses were reached through direct contacts. While for this chapter we cover 45 answers, the results we obtained still do not lend themselves well to statistical analysis. However, the results are still useful to:
\begin{itemize}
    \item identify characteristics of real-world problems that are not represented in artificial benchmarks, and
    \item identify real-world problems that could potentially be used as part of a benchmark suite.
\end{itemize}

From \todo{RP}\changed{8 April} 2020, an updated version of the questionnaire is available (see the end of Section~\ref{sec:outline} for details on the changes). In addition to addressing some of the shortcomings of the initial version, it has also been advertised more broadly (via mailing lists and newsletters)  therefore reaching a larger audience. This is the version presented in the following.

\subsection{Questionnaire Outline} \label{sec:outline}
\begin{figure}[t]
\centering
\includegraphics[width=.9\textwidth]{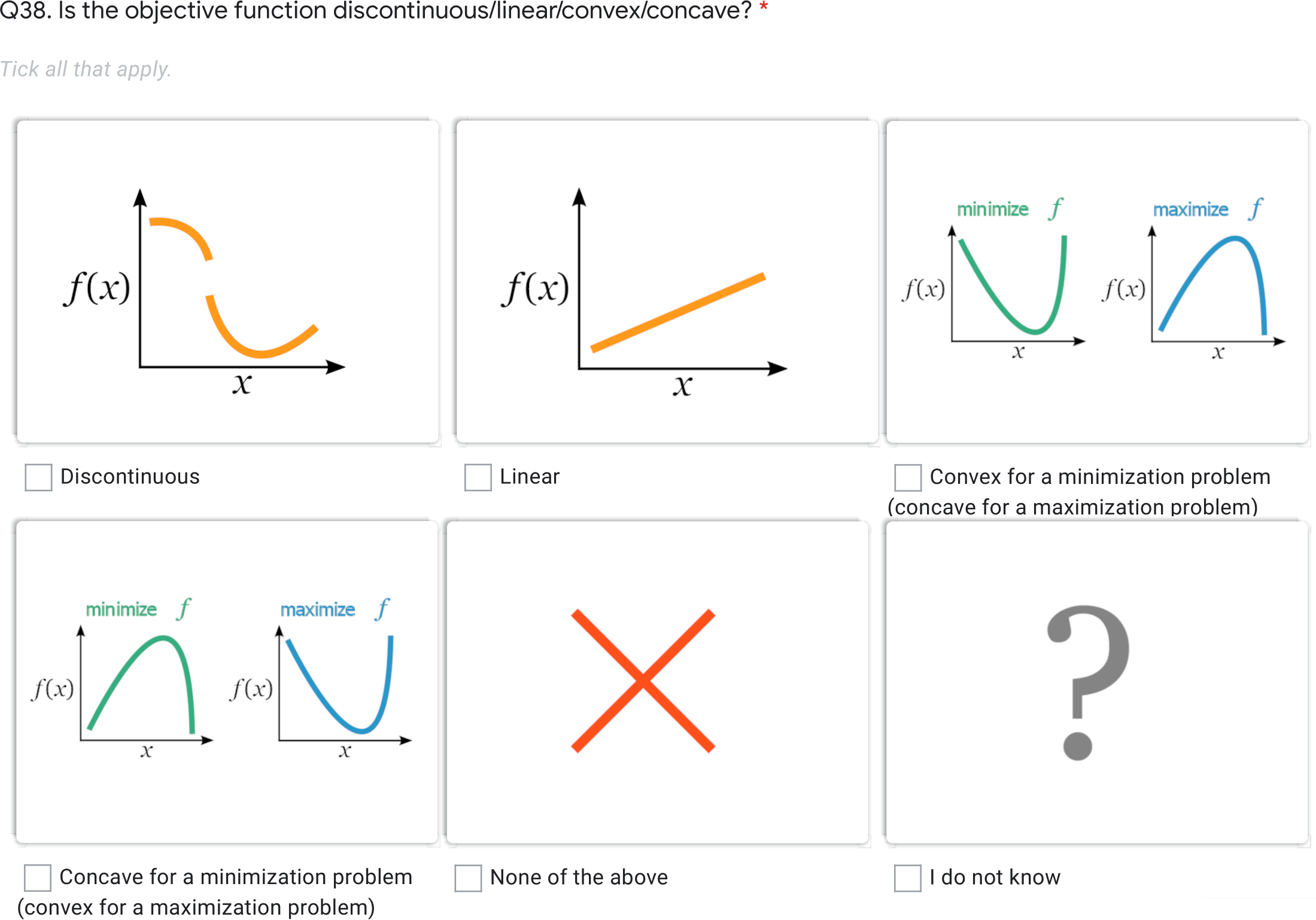}
\vspace{-0mm}
\caption{\todo{RP}\changed{Example question with concepts clarified by images.}}
\label{fig:question_example}
\end{figure}

The questionnaire aims to gather information on the types of issues that occur in real-world applications and how common the different issues (and their combinations) are. It consists of five parts, each addressing a different aspect of the problem (also see Figure~\ref{fig:survey_flow}):
\begin{figure}[t]
\centering
\includegraphics[width=\textwidth]{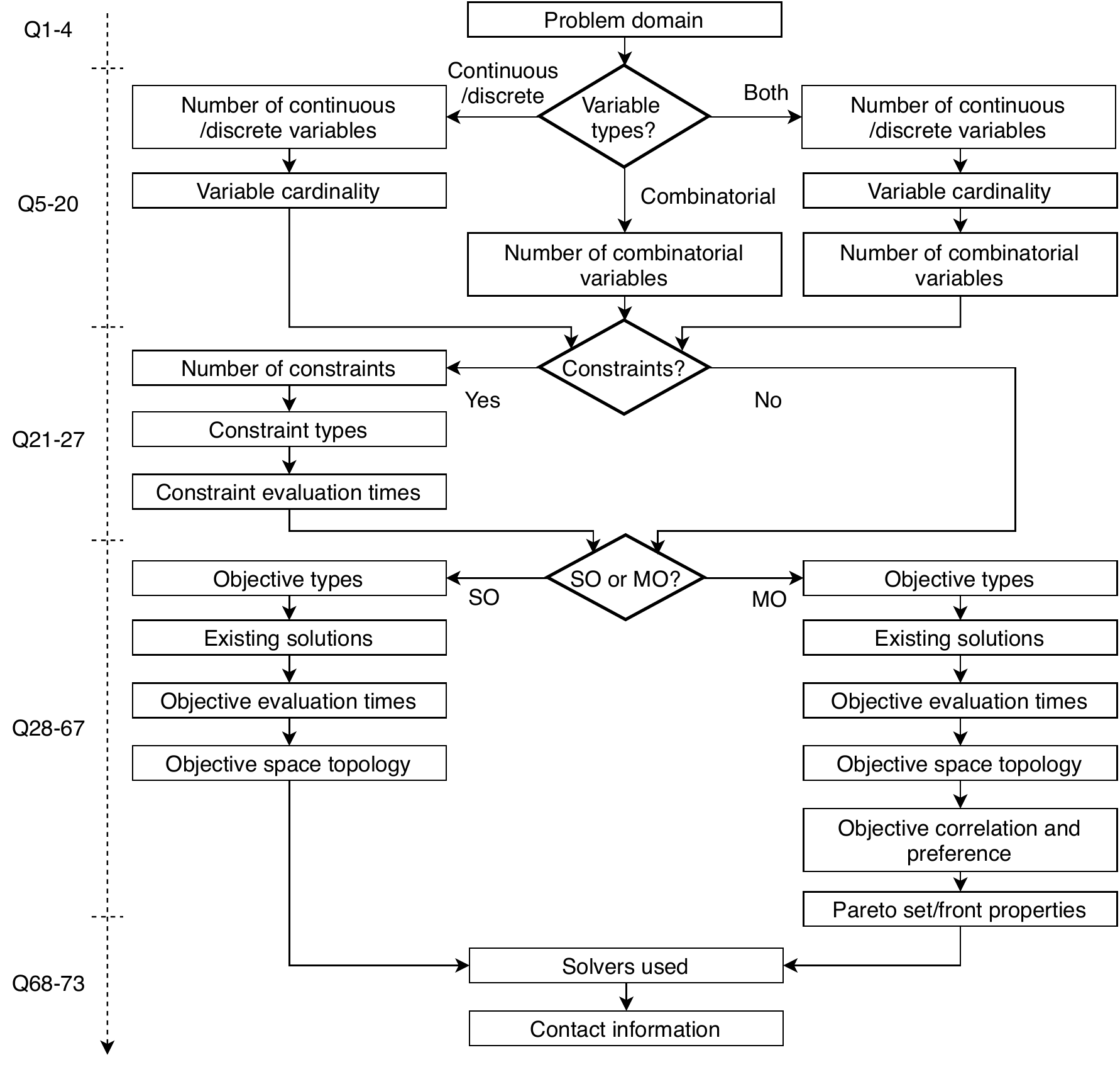}
\vspace{-4mm}
\caption{Questionnaire structure.
While there are 73 questions in total, depending on the problem properties, only between 28 and 51 questions are actually posed. 
}
\label{fig:survey_flow}
\end{figure}
\begin{itemize}
    \item  \textbf{Q1-4. General information} \\ 
    Four questions are devoted to the problem name, domain, existing sources with further details (such as papers) and whether any parts of the problem and/or its solutions are publicly available.
    \item \textbf{Q5-20. Questions about variables} \\ 
    \todo{R2-8}\changed{The questions about variables are split into three paths depending on the first question in this part which asks the type of variables present in the problem.} For each of the variable types (continuous, ordinal, categorical and combinatorial), the questionnaire asks the number of such variables, and 
    their cardinality where appropriate. 
    \item \textbf{Q21-27. Questions about constraints} \\ 
    The first question in this part asks whether the problem contains any constraints in addition to box constraints. 
    For constrained problems, further
    questions ask about several types of constraints (e.g. equality/inequality) and their evaluation time.
    \item \textbf{Q28-67. Questions about objectives} \\ 
    Here the first question asks whether the problem has a single objective, or multiple objectives (SO and MO in Figure~\ref{fig:survey_flow}, respectively). The remaining questions address the two cases separately. All 14 questions for SO problems are also asked for MO problems, but with adjusted phrasing (or are split into multiple parts).
    There are 26 questions in total for MO problems.
    \item \textbf{Q68-73 Final questions} \\ 
    The final six questions ask about the methods that were already used to solve this problem, contact information (unless the participant wishes to remain anonymous) and general comments regarding the questionnaire. 
\end{itemize}
In total, the questionnaire consists of 73 questions, but, depending on the problem properties, only between 28 and 51 questions need to be answered. 
\todo{???}\changed{The questions were formulated following terminology used in the Evolutionary Computation community as this is best understood by the authors and, presumably, by most of the respondents.} \todo{RP}\changed{In addition, for some questions images were used to clarify concepts as exemplified in Figure~\ref{fig:question_example}.}

The presented questionnaire is a modified version of the initial questionnaire. The changes were made based on the participants' feedback and our \changed{observations from the received results.} \todo{E1/E6} \changed{These changes were mostly made with the intention of striking a balance between the detail of the information gathered and the usability of the questionnaire, both in terms of clarity and the effort required to answer all questions. To condense the questionnaire further, we also removed questions where we did not expect a sufficient number of answers that could support quantitative analysis.} The changes we made comprise:

\begin{enumerate}
    \item adding a question about dynamically changing problem properties,
    \item removing questions about problem multidisciplinarity and explicit/implicit constraints,
    \item making a large part of the questions obligatory (and consequently adding the `I do not know' answer to obligatory questions), and 
    \item other minor adjustments in order to clarify some of the questions.
\end{enumerate}

Care was taken to make the modifications in such a way that answers to both versions of the questionnaire could still be used together. \todo{R2\_3} \changed{A plethora of additional questions were discussed as well, for example regarding hard- and soft constraints and multi-level problems. These potential changes were ultimately dropped in favour of keeping the questionnaire at a reasonable length.}

\section{Results}
\label{sec:results}

We have received 49 responses to the questionnaire by \todo{RP}\changed{16 July} 2020. Answers to both versions of the questionnaire were reconciled by, whenever possible, updating the earlier responses via private correspondence with respondents or referring to literature provided in the responses. After verification, some of the responses had to be discarded (due to duplication and errors), resulting in a total of 45 verified responses. References to literature submitted by the participants \todo{RP}\changed{and the data used in the analysis} are publicly available online \cite{BloDeiMarNau2020}.

\begin{figure}[t]
\centering
\includegraphics[scale=0.65]{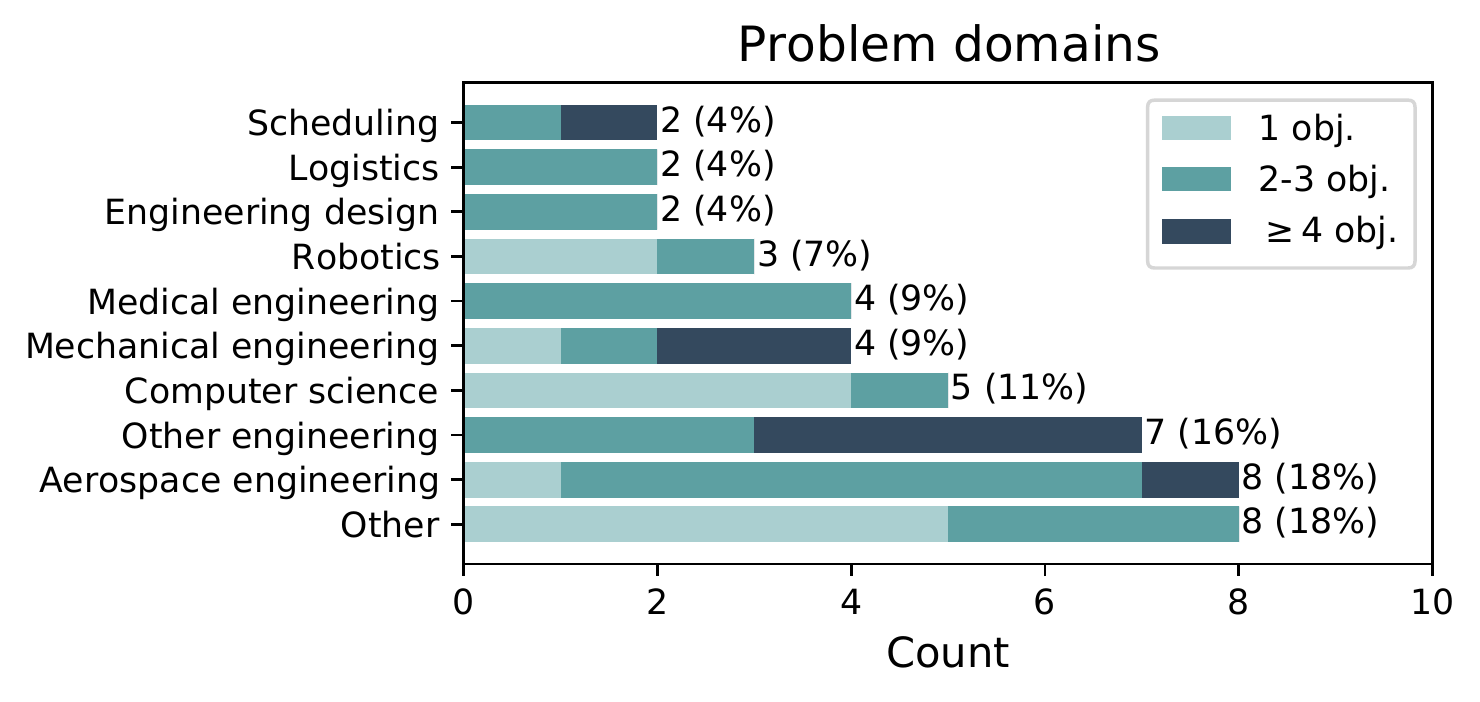}
\vspace{-4mm}
\caption{Number of problems per domain. Based on 45 responses. Domains merged into Other: Bioinformatics, Chemistry, Energy, Manufacturing, Natural hazard assessment, Operations research, Photonics, Semiconductors}
\label{fig:problem_domain}
\end{figure}

\begin{figure}[t]
\centering
\includegraphics[scale=0.65]{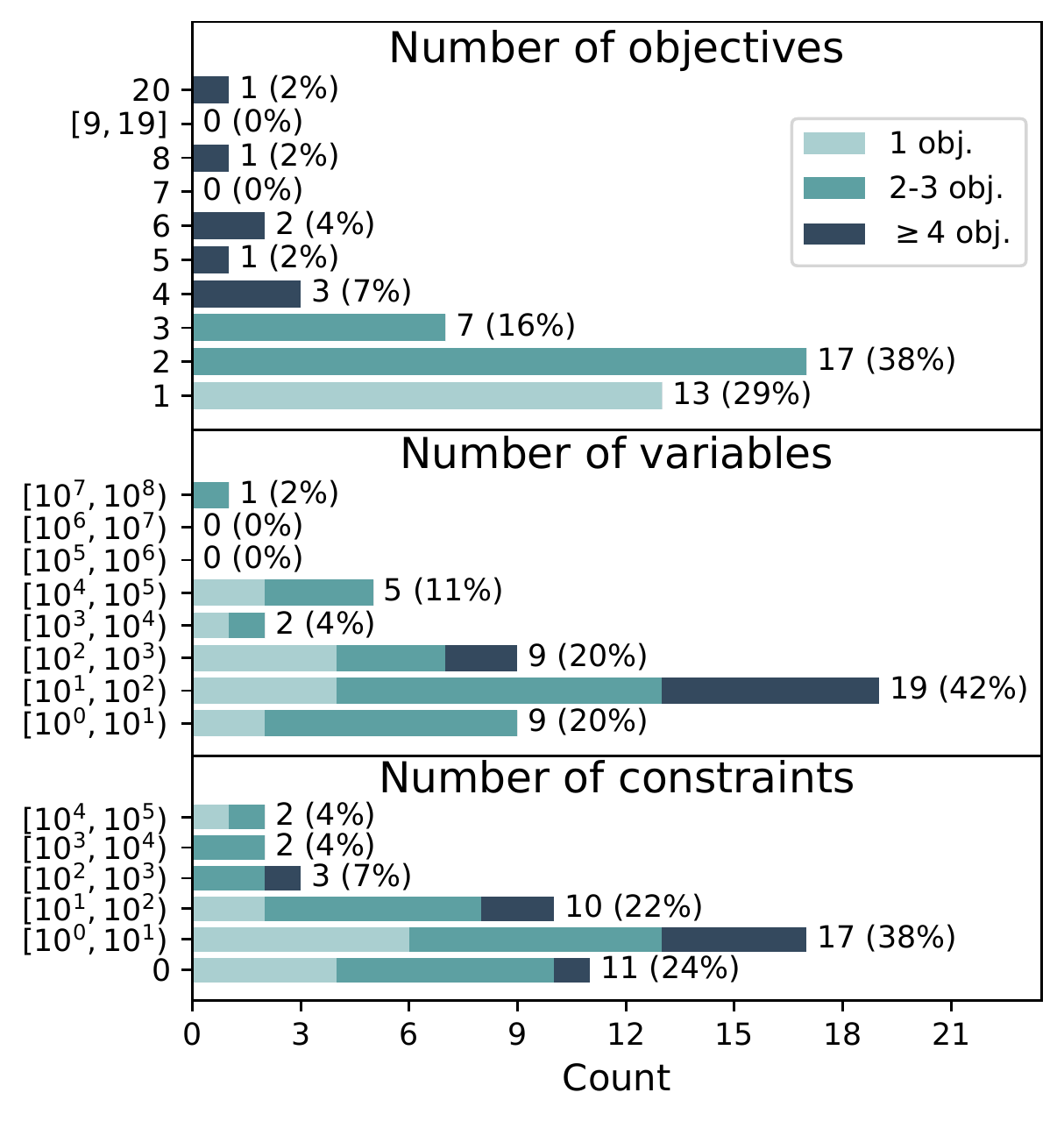}
\vspace{-4mm}
\caption{Problem frequency per number of objectives (top), variables (middle) and constraints (bottom). Based on 45 responses.}
\label{fig:numbers}
\end{figure}

\begin{figure}[t]
\centering
\includegraphics[scale=0.65]{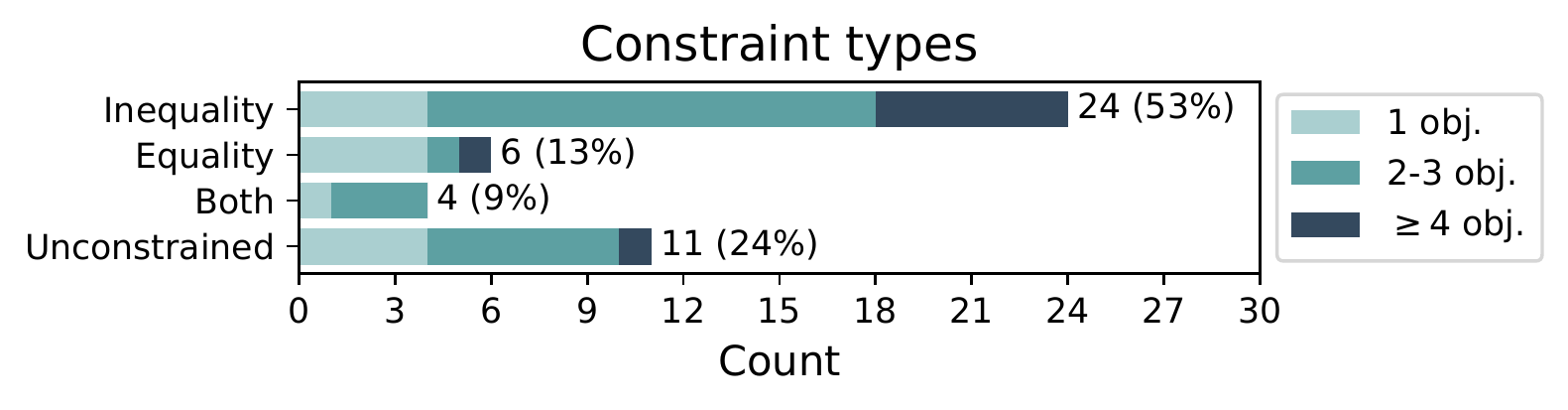}
\vspace{-4mm}
\caption{Number of problems with equality, inequality, both types of constraints, or no constraints. Based on 45 responses. }
\label{fig:equality_inequality_constraints}
\end{figure}

\begin{figure}[t]
\centering
\includegraphics[scale=0.65]{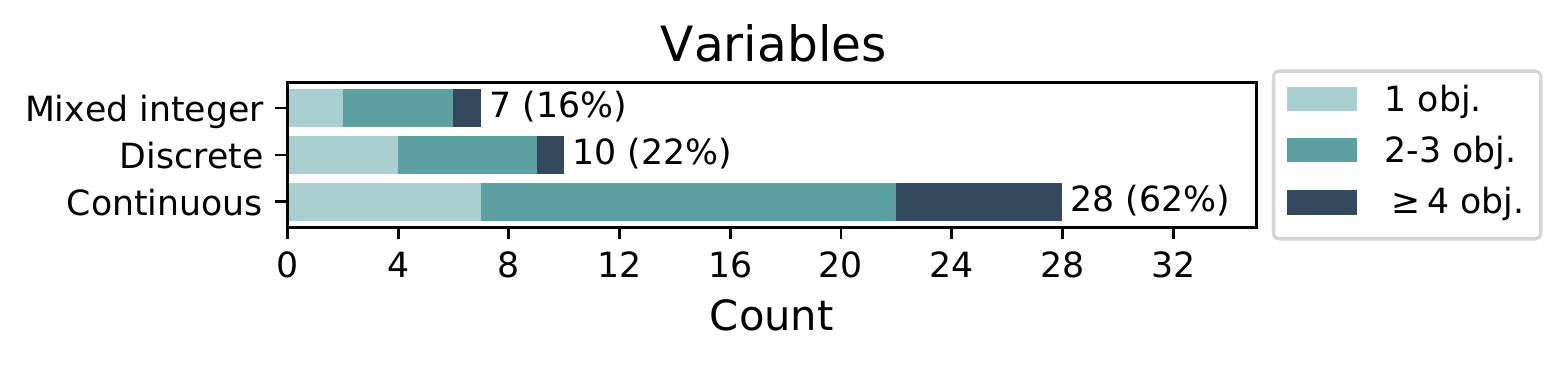}
\vspace{-4mm}
\caption{Frequency of different combinations of variable types. Based on 45 responses.}
\label{fig:variable_types}
\end{figure}

\begin{figure}[t]
\centering
\includegraphics[scale=0.65]{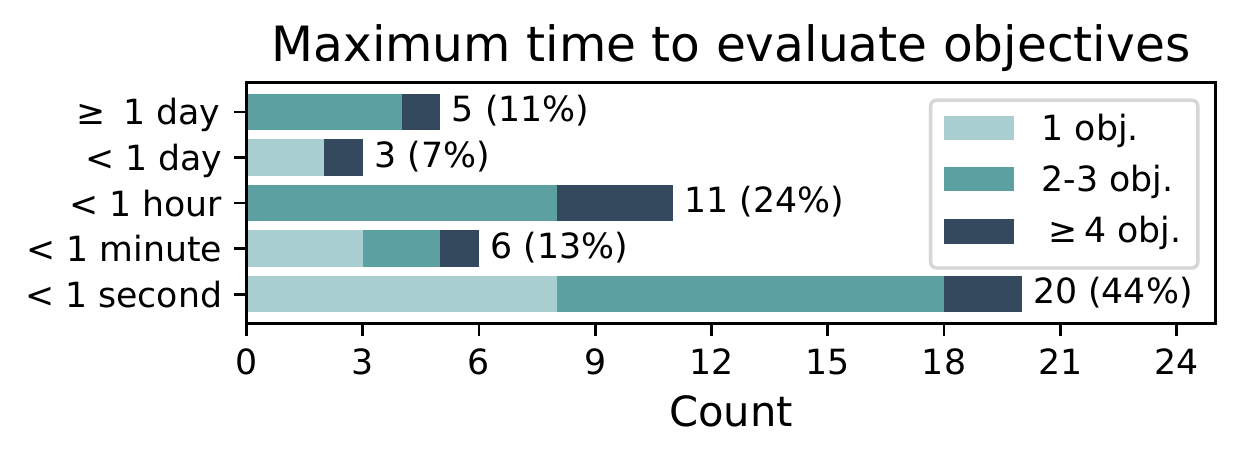}
\includegraphics[scale=0.65]{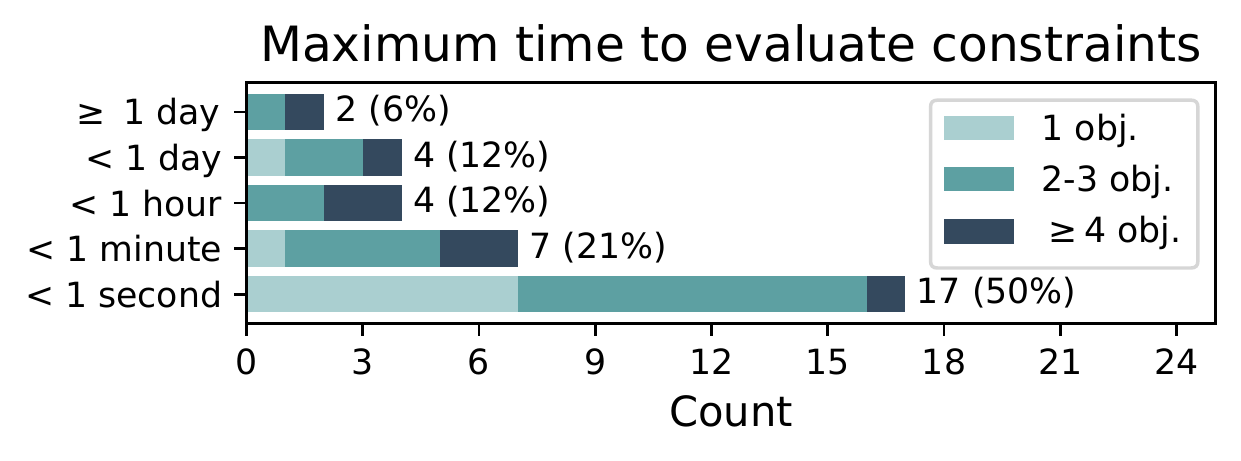}
\vspace{-4mm}
\caption{Top: Frequency of evaluation time ranges for objective functions, based on 45 responses. Bottom: Frequency of evaluation time ranges for constraints, based on 34 constrained problems.}
\label{fig:evaluation_times}
\end{figure}

\begin{figure}[t]
\centering
\includegraphics[scale=0.65]{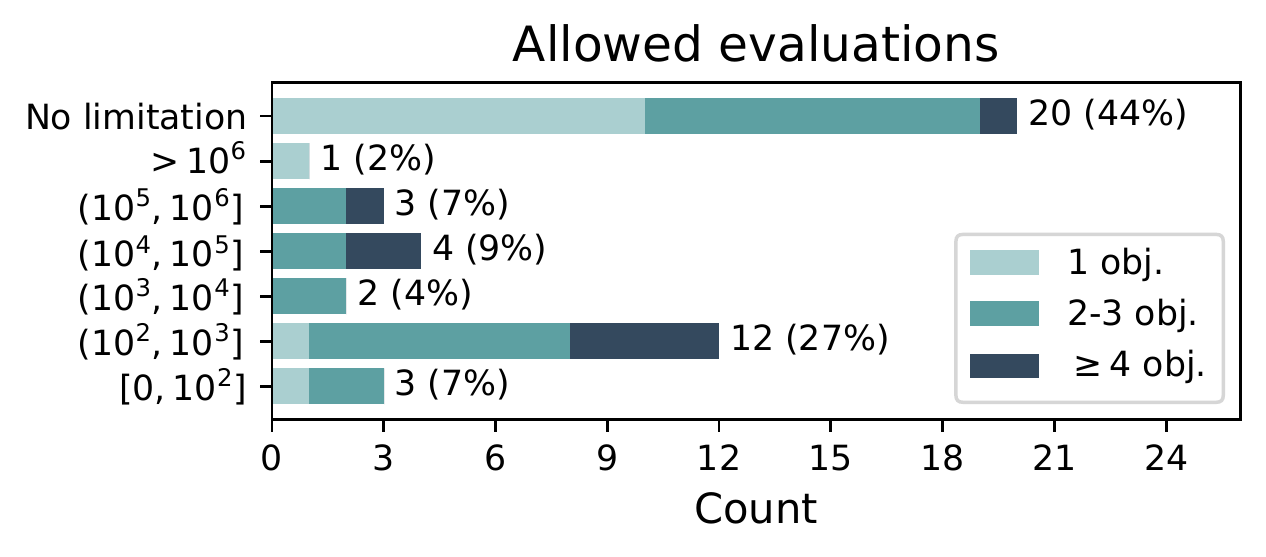}
\vspace{-4mm}
\caption{Number of evaluations that can be used. Based on 45 responses.}
\label{fig:allowed_evaluations}
\end{figure}

\begin{figure*}[t]
\centering
\includegraphics[width=\textwidth,trim={0pt 15pt 45pt 0pt},clip]{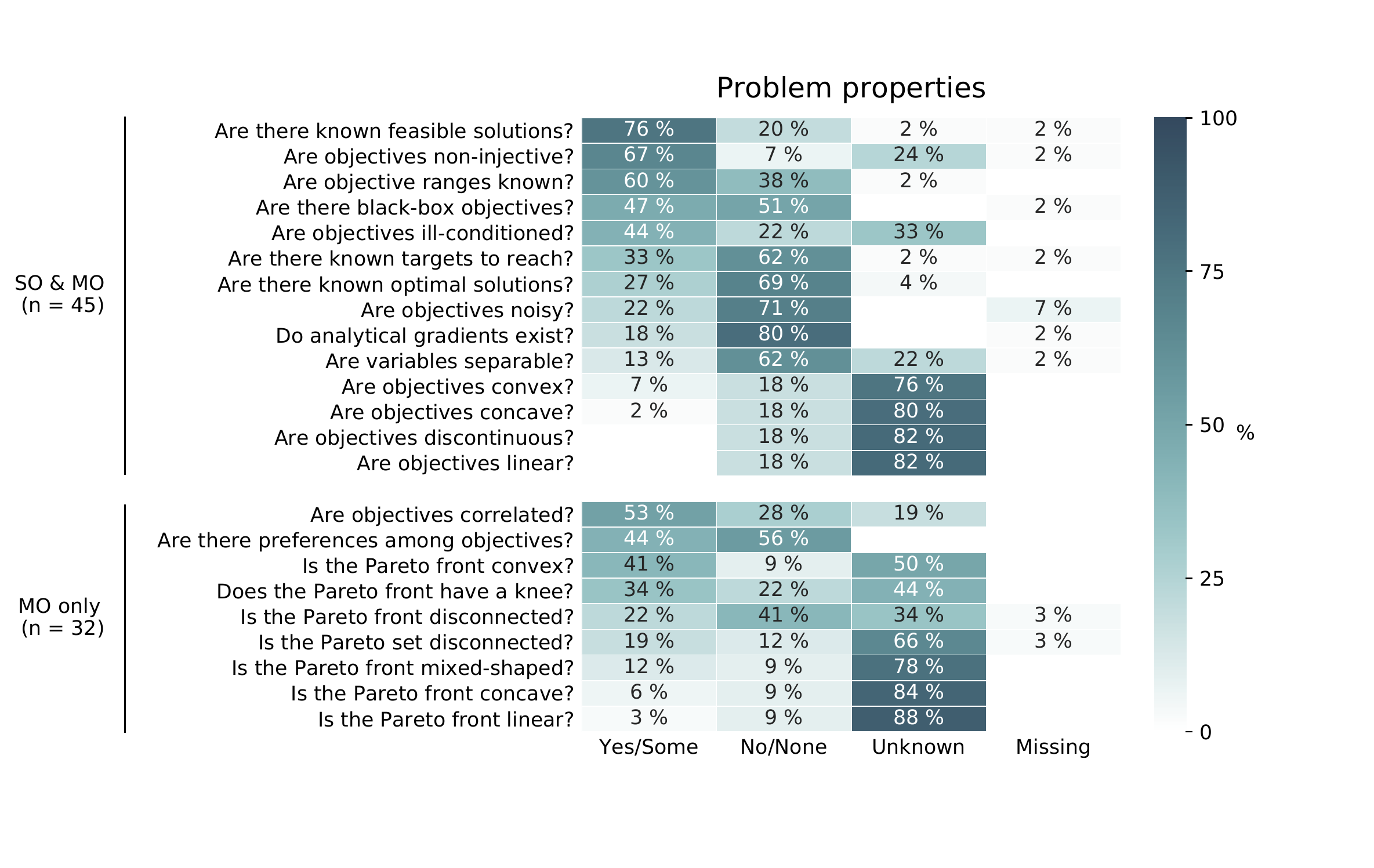}
\vspace{-4mm}
\caption{Heatmap showing the percentage of various problem properties both for questions shared by single-objective (SO) and multi-objective (MO) problems (45 responses), as well as for questions specific to MO problems (32 responses). }
\label{fig:objective_properties}
\end{figure*}

\begin{figure}[t]
\centering
\includegraphics[scale=0.65]{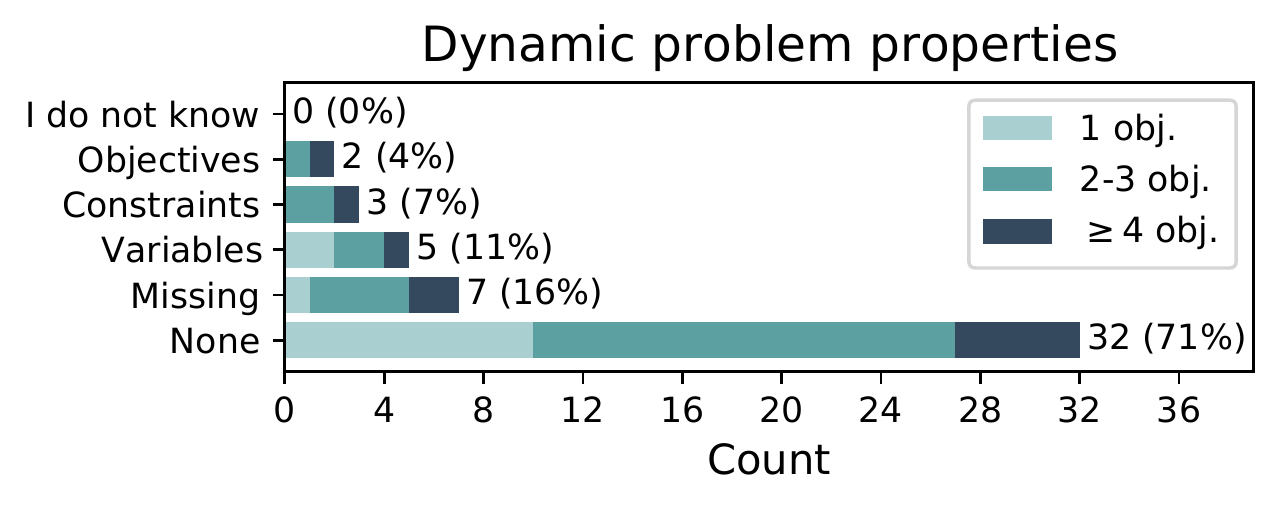}
\vspace{-4mm}
\caption{Frequency of dynamic problem properties. Percentages are given w.r.t. the total number of participants. Based on 45 responses.}
\label{fig:dynamic_properties}
\end{figure}

\begin{figure}[t]
\centering
\includegraphics[scale=0.65]{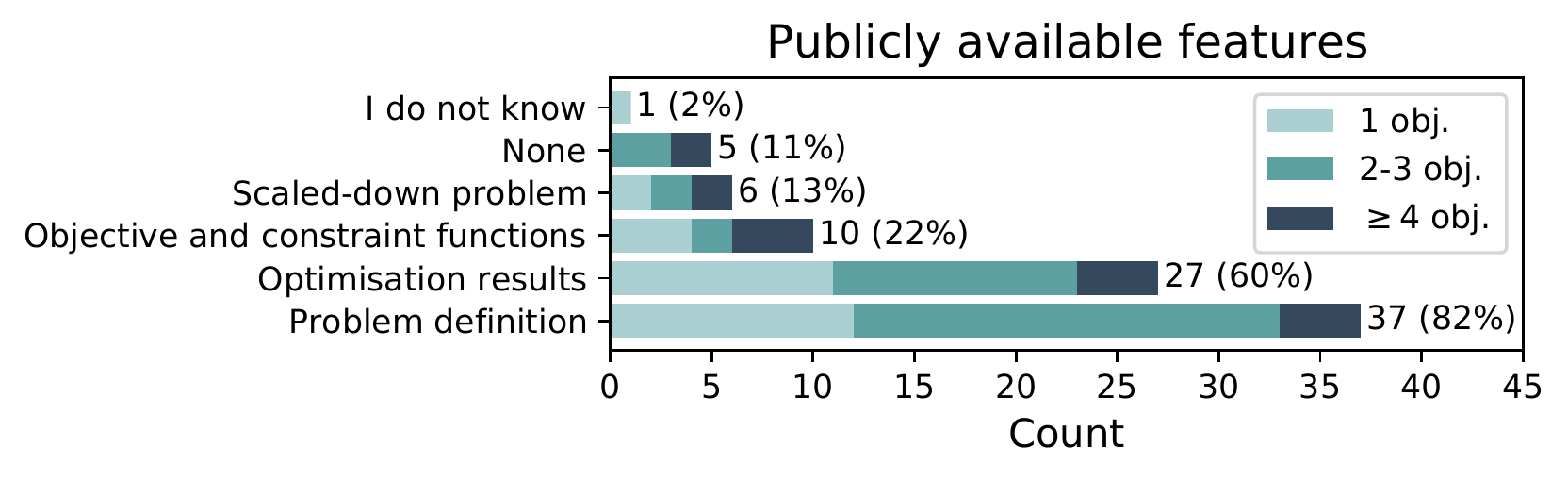}
\vspace{-4mm}
\caption{Frequency of publicly available features. A selection of the multiple responses per problem. Percentages are given w.r.t. the total number of participants. Based on 45 responses.}
\label{fig:availability}
\end{figure}

The results are presented with bar plots 
(Figures \ref{fig:problem_domain} to 
\ref{fig:allowed_evaluations}, \todo{E-7}\changed{\ref{fig:dynamic_properties} and \ref{fig:availability}}) 
and a heatmap (Figure \ref{fig:objective_properties}). In all bar plots, the $x$-axis indicates the number of problems featuring the given property. Relative frequencies are provided next to each bar. Colours are used to differentiate between single-objective, multi-objective (bi- and tri-objective), and many-objective (four or more objectives) problems following common practice in evolutionary multi-objective optimisation.

Within the questionnaire, we ask participants for the domains of their optimisation problems. The results are summarised in Figure \ref{fig:problem_domain}. As expected, most of the submitted problems are engineering problems (25 out of 45), spread over several sub-fields. However, we also received problems from applications in computer science, robotics, logistics, scheduling, bioinformatics, chemistry, energy, manufacturing, natural hazard assessment, operations research, photonics and semiconductors. This suggests that real-world optimisation problems can be found in a wide range of applications.

This result means that investigating domains beyond engineering could yield more and different examples of problems for a real-world benchmark of evolutionary algorithms. At the same time, this also indicates that potential benchmarking problems could have very differing characteristics, based on their respective application. This should be kept in mind for the analysis below, but also in general when compiling real-world benchmarks.

Figure \ref{fig:numbers}
summarises the results in terms of the number of objectives (top), variables (middle) and constraints (bottom). 
Most problems feature two objectives (17 responses), followed by problems with one objective (13 responses) and three objectives  (seven responses). The eight many-objective problems have between four and 20 objectives.
These results suggest that researchers in our community are mainly faced with multi- and many-objective problems. \todo{R1-1}\changed {This finding is likely due to biased sampling in a community disproportionately concerned with multi- and many-objective optimisation.} However, as demonstrated in \todo{R2-11}\changed{Chapter} \ref{chap:bench}, multi- and many-objective benchmarks are still rare. We thus hope to encourage colleagues to release and share more problems and benchmarks in the future. First efforts towards this direction have already begun with the benchmarking network but the collected 
resources are still far from comprehensive (cf. \cite{BenchNet2019}).

The high number of objectives further suggests that the standard assumption of conflicting objectives might not hold for some problems, at least not across the entire search space. Out of the eight problems submitted to the questionnaire with four or more objectives, half of them feature some kind of correlation between objectives. Two others did not have correlations (two of the four-objective problems), and for the last two problems it is not known whether there are correlations (one of the six-objective problems and the 20-objective problem). Out of the bi- and tri-objective problems thirteen have some kind of correlation, seven have no correlation, and for the other four the existence of correlations is unknown. This means that in total just over half of the multi- and many-objective problems submitted to the questionnaire involve correlations. In \todo{R2-11}\changed{Chapter} \ref{chap:correlation} correlations between objectives are discussed further, but these are as of yet not often considered in benchmarking. This should be remedied to enable research into algorithms that efficiently handle the resulting dependency structures.


The middle part of Figure \ref{fig:numbers}  conveys the number of variables of the collected problems in powers of ten.
Most problems possess between 10 and 100 variables (19 responses), followed by nine responses with less than ten variables, and nine responses with between 100 and 1000 variables. While only \SI{20}{\percent} of the problems have a small search space with ten variables or less, and benchmarks are scalable beyond that, in practice this is not always done. Based on this data it seems to be good practice to scale benchmark problems to higher dimensions, although this may come at a computational cost.
While the largest problem in terms of the number of variables contains $10^7$ (ordinal) variables, the other problems contain less than $10^5$ variables. Almost \SI{40}{\percent} (17 out of 45) of the problems have 100 variables or more, something rarely seen in benchmarks. To reflect the dimensionality of real-life problems in artificial benchmarks, analysis might be done to see whether larger problems are needed, or if scaling up existing benchmarks would be sufficiently accurate. \todo{E8}\changed{This also seems to apply to many-objective optimisation, considering that two of the eight many-objective problems also have more than 100 variables.}

The bottom part of Figure \ref{fig:numbers} highlights the number of constraints (other than box constraints).
There are eleven unconstrained problems, and 27 with less than 100 constraints. 
The largest problem\todo{grammar}\changed{,} in terms of constraints\todo{grammar}\changed{,} consists of \num{45000} constraints (one equality and the rest inequality constraints).
Overall it is clear that constraints are very common in \todo{E8}\changed{single-, multi- and many-objective problems,} and quite often appear in large numbers. Benchmarks with constraints are definitely needed. Further, it may be worthwhile to create a taxonomy of different constraint types to enable better analysis. This would be akin to fitness landscape descriptors (e.g. multi-modality), but then specifically focusing on constraints. It would include, for example, the balance between equality and inequality constraints, and the ratio of feasible solutions.
 
The findings of Figure~\ref{fig:numbers}
are in line with the study on industrial optimisation problems in \cite{TiwNorHutTur2015}, where a large number of variables and constraints and multiple objectives are found.

Based on Figure~\ref{fig:equality_inequality_constraints}, inequality constraints occur more frequently than equality constraints (in 24 versus six problems, while four have both types of constraints). Notably, \todo{RP}\changed{based on additional data from the results website \cite{BloDeiMarNau2020},} for almost half of the constrained problems (15 responses), evaluation of the objectives is not always possible when constraints are violated. \todo{R2\_2} \changed{This is an instructive finding as solutions that cannot be evaluated are not commonly modelled in most popular benchmarks.}

Figure \ref{fig:variable_types} provides the frequency of different variable types. Most submitted problems contain only continuous variables (28 responses), followed by problems with only discrete variables (ten responses) and mixed-integer problems (seven responses). The prevalence of continuous problems holds also when the results are broken down with regard to the number of objectives. The presence of ordinal, categorical and permutational variables (not shown in the figure) is rather balanced. The number of non-continuous problems containing ordinal, categorical and permutational variables are eight, eight and seven, respectively, with five problems comprising two or three of these variable types. 

While the portion of mixed-integer problems in the questionnaire responses is significant, research about mixed-integer problems in Evolutionary Computation is limited. Some mixed-integer benchmarks are available, but\changed{,} as discussed in \cite{TusBroHan2019gecco}\changed{,} the existing mixed-integer linear programming problems such as \cite{BusDruMee2003} are often hard to use for black-box optimisers like evolutionary algorithms. 
Because integers can be used to represent ordinal, categorical and permutational variables, most mixed-integer problems make no distinction among them. For example, \cite{McCKee2011cec,SadThiBos2018ec} consider only continuous and binary variables, while \cite{TusBroHan2019gecco} is limited to continuous and integer variables. A few \todo{R2-9}\changed{(like \cite{LiaSocdeOStu2013tec})} comprise continuous, ordinal and categorical variables, even less contain a mix of permutational and other variables (e.g. \cite{PolyBonyWagnMichNeum2014}). 
Single and bi-objective mixed-integer problems were proposed in \cite{TusBroHan2019gecco}, while more generic multi-objective (scalable in the number of objectives) problems were introduced in \cite{McCKee2011cec}. It is important to highlight these benchmark suites as quite frequently authors propose their own, new problems, rather than using an established well \changed{thought-out} suite. 

Information about the maximal cardinality of ordinal and categorical variables, as well as about discrete objectives is limited because each of these properties occurred in at most \SI{20}{\percent} of the responses. However, a few observations can be made. The six problems with ordinal variables are scattered over various categories \changed{of cardinality}, ranging from binary to more than \num{10000}. Unsurprisingly, for the categorical variables the cardinality is always low with the highest being in the 11 to \num{100} interval. Nine problems with discrete objectives were submitted. Most of these have a high cardinality of more than \num{10000}, with the remainder being distributed over a variety of lower cardinalities.

Real-world optimisation problems sometimes exhibit time-consuming function evaluations.
Figure \ref{fig:evaluation_times} reports on maximum evaluation times of the objective functions for all problems (top plot) as well as of the constraint evaluation times for constrained problems (bottom plot). We can see that \changed{most constrained problems} have a low maximum evaluation time, although \SI{18}{\percent} exhibit very long maximum evaluation times, i.e. one hour or more. A similar distribution of maximal evaluation times can be observed for objectives, however, not as monotone.
Again, in \SI{18}{\percent} of the problems it takes an hour or more to evaluate the most expensive objective function. Note that while single-objective problems seem to have shorter evaluation times than multi- and many-objective ones, this could be an artefact of showing only maximum evaluation times. These are more likely to be long if more objectives are considered.
Unsurprisingly (based on the raw data), the time taken to evaluate constraints and the time taken to evaluate objectives are in the same range for most problems, although there are some exceptions.

For both objectives and constraints, there are quite a few problems with short evaluation times. With these real-world problems, a benchmark suite could be compiled that would also be practical in terms of evaluation time. This would be particularly interesting if these problems have properties that are not covered by real-world benchmark suites like RE and CRE \cite{TanIsh2020asc}.
On the other hand, a lot of problems with relatively expensive evaluations (ranging from minutes to days) were also submitted to the questionnaire. This indicates that handling expensive evaluations is another important research direction. It could be investigated whether problems with expensive evaluations are inherently different. When problems are based on simulations, they might have different constraint structures. In that case, it may be possible to replicate those constraint structures in artificial, and cheap, functions. Otherwise, if they do not have different properties, existing benchmark functions could be used with a small evaluation budget (e.g. $<100$). These could be other real-world problems with relatively cheap simulations like those from the GBEA \cite{VolNauKerTus2019gecco} or CFD \cite{DanRahEveTab2018ppsn} suites.

In addition to the evaluation time of objectives and constraints, we also asked how many solution evaluations were `allowed' for each problem, the results of this are shown in Figure~\ref{fig:allowed_evaluations}. A lot of problems have no limitation on the number of evaluations, which matches existing benchmarking practice. On the other hand, a large number of problems is severely restricted in the number of evaluations that can be used. There seems to be a particularly great need for algorithms that can optimise while restricted to at most \num{1000} function evaluations. In the limited data available, we observed a clear correlation between the number of evaluations and the evaluation time. A small number of evaluations is allowed for the problems which need long evaluation times. Of the four most time-consuming problems that take more than one day for the evaluation of objectives and constraints, two allow for less than 100 evaluations and two for less than 1000 evaluations.

The heatmap in Figure \ref{fig:objective_properties}
provides the frequency of
answers to a selection of the questions.
Here, questions are grouped vertically based on whether they apply to single- as well as multi- and many-objective problems (SO \& MO), or only to multi- and many-objective ones (MO only). In addition, the questions in these two groups are sorted based on the frequency of 'Yes/Some' answers.
The majority of responses state that there are no known optimal solutions for their problems. 
Therefore, directly integrating real-world problems into benchmark suites would complicate performance measurements as some metrics would not be applicable, e.g., distance to optima or the (inverse) generational distance \cite{CoeCor2005,VelLam2000cec} to the Pareto front. 
Furthermore, in many cases there are no known targets that could be exploited by algorithms for guidance or to terminate optimisation runs.  Analytic gradients are not available in most problems either. This limits the use of gradient-based optimisation algorithms. 
Realistic benchmarks should preferably reflect these limitations.

Contrariwise, real-world problems are reported to exhibit characteristics that could potentially be exploited in algorithm design and, therefore, should be emulated in benchmarks. For example,
\begin{itemize}
    \item objectives are often reported to be correlated,
    \item feasible solutions are known for most problems, which simplifies search initialisation,
    \item objectives are predominantly deterministic,
    \item the presence and absence of black-box objectives is fairly balanced, 
    \item preferences among objectives exist in about half the problems.
\end{itemize}
The answers also reveal that characteristics of real-world problems are not fully understood: participants often could not say whether their Pareto sets are disconnected, Pareto fronts are linear, concave, convex, mixed-shaped or
have knee points, nor whether objectives are
convex, concave, discontinuous, or linear.
Designing artificial benchmarks solely based on empirical observations from real-world problems might produce problems of too low complexity. In contrast, directly reusing real-world problems as benchmarks would overcome this shortcoming. Alternatively, systematic automated analyses of existing real-world problems could reveal and collect properties that could not be gathered in questionnaires. This may be done by exploratory landscape analysis for example \cite{MersBiscTrauPreuWeih2011}.

\todo{E-7 (swapped this and the next paragraph)}
Frequencies of dynamic problem properties are shown in Figure~\ref{fig:dynamic_properties}. The majority of problems (32 out of 45) have no dynamic properties, while a few problems contain dynamic variables, constraints, and objectives. This information is missing for seven problems out of 45. Despite relatively few dynamic problems having been submitted, it is still a significant portion. This observation supports the need for continued research in this area, including the development of benchmarks with dynamic properties. 

The bar plot in Figure~\ref{fig:availability} provides the frequencies of publicly available problem features such as problem definition, optimised results, objective and constraint function evaluation modules and a scaled-down version of the problem.
Only six problems have no publicly available features (or their availability is unknown), while most of the problems (37 out of 45) have at least a publicly available problem definition.
More than half (27 out of 45) of the responses stated that optimisation results are publicly available.
Considering that for many problems at least one feature is available, it is unfortunate that in many cases this is not exploited. It is likely that if the problems were gathered in an easily executable and stable suite, these real-world problems would be much more widely studied. An issue with real-world problems that might complicate this is that they may have platform specific requirements, particularly in the case of simulation software. A tool like Docker\footnote{\url{https://www.docker.com}} might help to ensure cross-platform execution of a benchmarking suite that includes these problems. In addition to the existence of a benchmark suite, its advertisement is also important to increase its reach. This can be achieved in part by sharing the benchmark suite through initiatives such as the benchmarking network (cf. \cite{BenchNet2019}).

\section{\changed{Conclusions}}
\label{sec:conclusion}

Much research exists that empirically compares the performance of different algorithms using benchmarks.
This work is motivated by one of the primary issues of benchmarks: the lack of clarity of whether and how conclusions drawn from benchmark results can be applied to optimisation problems in real-world applications. When it is not clear what the performance of an algorithm on a benchmark means for the performance on a real-world problem, it becomes difficult to choose a good optimisation algorithm for a given problem.

To mitigate this issue, benchmarks should be created that contain problems that closely resemble real-world problems. In order to effectively design such test problems,
the properties and characteristics of real-world problems have to be understood. To this end, this chapter proposed a questionnaire designed to elicit information on real-world problems from \changed{industrial and academic optimisation experts.}

\todo{E8 -- New subsection}\changed{\subsection{Discussion}}


\changed{The current set of 45 responses is limited, and not randomly sampled from the optimisation community, and therefore precludes generalisation. Even so, these initial findings already reveal apparent gaps in currently available benchmark suites. 
The obtained results indicate that many real-world problems are continuous and constrained. Not only are constraints very common, they quite frequently appear in large numbers. Both of these observations indicate a need for benchmarks focused on constraint handling (also see \cite{TanOya2017cec}).} There are clear trends in objective function properties, such as a prevalence of deterministic objectives, and about half of the analysed problems have black-box objectives. All of these observed properties should be reflected in benchmark design. Some findings also pose challenges to choosing benchmark design strategies: for many of the real-world problems submitted to the questionnaire the respondents indicated that the problem properties (e.g. the shape of the Pareto front) are unknown. Thus, designing faithful artificial benchmarks is error-prone.

Based on the results obtained in our study, there are several real-world problems with evaluation times that are low enough (less than one second) to potentially result in practical computational costs when compiled into a benchmarking suite. However, it is unclear whether these problems are also able to cover a wide enough range of problems to create suitable diversity for a general benchmark. Further research on the properties of real-world problems and how to create similar artificial problems is therefore still needed.

An additional issue that is not solved by compiling real-world problems into a benchmark directly is the reliance on reference solution sets for many performance measures. While we found that several authors published their optimisation results which could be used as a stand-in, the results would be biased since these solutions are likely not optimal. Besides additional investigation of performance measures (see \todo{R2-11}\changed{Chapter} \ref{chap:measures}) it will thus still remain worthwhile to identify the properties of real-world problems and create artificial functions with known optima accordingly.

Some other issues may also be addressed when using such artificial functions. Many problems were indicated to have severe restrictions on the number of evaluations ($< 1000$). This \todo{style/grammar}\changed{finding suggests that} research is needed in this area, particularly since currently there is not much attention to this. When benchmark functions are evaluated with such restrictions in mind\todo{grammar}\changed{,} they may still give an indication about performance with a limited budget.

Additional observations are that problems come from diverse domains, and are not limited to the relatively well covered field of engineering. In turn, these diverse problems may also have diverse characteristics, which should be reflected in benchmarks. Further, almost \SI{40}{\percent} of the submitted problems \todo{style}\changed{have} 100 variables or more. To cover this in benchmarks, larger problems are needed. Alternatively, existing benchmarks might be scaled up, as long as it is confirmed that this is a sufficiently \todo{grammar}\changed{accurate} representation of high-dimensional real-world problems.

Since more than half of the gathered multi- and many-objective problems include correlated objectives, benchmarks are needed in this area as well (also see \todo{R2-11}\changed{Chapter} \ref{chap:correlation}). Mixed-integer problems do not appear to be very common, but still represent a significant fraction of the responses. Although some mixed-integer benchmarking problems exist, there also still appear to be some gaps such as a suite including continuous, integer, and categorical variables in a multi- and many-objective setting. Finally, benchmarks are needed that are dynamic in their objectives, constraints, and/or variables.

\todo{E8 -- New subsection}\changed{\subsection{Highlights for many-objective optimisation}}

\todo{E8}\changed{
In line with the focus of this book, i.e. many-objective optimisation, we highlight some findings that are especially relevant for the many-objective case. Although the questionnaire covered all of single-, multi-, and many-objective optimisation, a significant part of the responses (8 out of 45) concerned many-objective problems.
}

\todo{E8}\changed{
Overall, \SI{53}{\percent} of the multi- and many-objective problems involved correlated objectives, which carries over to the many-objective problems where this was exactly \SI{50}{\percent}. Correlations are especially relevant for many-objective problems, because detecting them may make it possible to reduce the number of objectives. Naturally, with a lower number of objectives the problem becomes more manageable.
In turn, techniques from multi-objective optimisation might even become applicable in some cases.
}

\todo{E8}\changed{
Similarly, preferences among objectives may also be used to reduce the number of objectives. By focusing on the most preferred objectives first, the number of objectives handled at the same time is reduced. 
With almost half of the gathered multi- and many-objective problems (\SI{44}{\percent}) having preferences among objectives, this is clearly an important area for research in many-objective optimisation.
}
%

\todo{E8}\changed{
A problem that seems to be even more relevant in the many-objective case is the length of evaluation times for objectives. With a larger number of objectives, the probability of any one of them being expensive to evaluate grows. This is also supported by the questionnaire responses, where the (maximum) evaluation time for the objectives was shorter for single-objective problems compared to multi- and many-objective problems.
}

\todo{E8}\changed{
Results drawn over the full set of responses often also apply to the many-objective subset. For instance, this is the case for problems with more than 100 variables, the prevalence of constraints, and the prevalence of continuous problems. So in addition to generally needing more attention and inclusion in benchmarks, these issues also need to be addressed in the many-objective case.
}

\todo{E8 -- New subsection}\changed{\subsection{Future Work}}

Based on our observations, there are several big open tasks for the future:
\begin{itemize}
    \item Identify problems that would be suitable for compilation into a benchmark based on their computational costs and implementation ease, as well as property coverage.
    \item Run a data-driven analysis of the properties of real-world problems (for example using exploratory landscape analysis \cite{MersBiscTrauPreuWeih2011}) where solution sets are available to gather additional information on the problems and identify potential trends.
    \item Survey existing benchmarks with artificial functions and compare the properties that are represented. Develop new artificial functions (ideally with known optima) with the missing characteristics.
\end{itemize}
\todo{E2,E3,E4}\changed{Furthermore, the questionnaire does not collect information about several relevant problem characteristics: the uncertainty in objective and constraint evaluation, whether problem descriptions have multiple representations in which objectives can become constraints (and vice versa) dependent on the current solution, whether problems are real-time, etc. Investigating such aspects may have merit.}
In addition, we would like to be able to do a quantitative analysis of our \changed{questionnaire} results. While we were already able to make some interesting observations, 45 responses are not enough to compute reliable statistics.
We thus hope to collect further data in order to be able to
strengthen the understanding of real-world problems and discover more intricate patterns in the data. The reader is thus encouraged to fill out and share the questionnaire that can be found at: \url{https://tinyurl.com/opt-survey}.

\section*{Acknowledgments}

The questionnaire was first proposed at the Lorentz Center MACODA (Many Criteria Optimization and Decision Analysis) workshop as a group effort by 
J.\ Fieldsend, 
J.\ Forde, 
H.\ Ishibuchi, 
E.\ Marescaux, 
M.\ Miyakawa, 
R.\ Purshouse, 
J.\ Richter,
D.\ Thierens,
C.\ Tour{\'e},
and the authors of this paper. We thank the working group participants for their valuable contributions.
We also wish to thank 
R.\ Allmendiger, 
J.\ Alza-Santos,
M.\ M.\ Awad, 
M.\ Balvert, 
L.\ Bliek, 
A.\ Bouter, 
B.\ Breiderhoff, 
K.\ Chiba, 
C.\ Doerr, 
M.\ Ehrgott, 
M.\ Erascu, 
D.\ Gaudrie, 
M.\ Interciso, 
M.\ Kanazaki, 
J.\ Knowles, 
T.\ Kohira, 
P.\ Z.\ Korondi, 
O.\ Krause, 
W.\ B.\ Langdon, 
M.\ van der Meer, 
N.\ Namura, 
M.\ Ohki, 
Y.\ Ohta, 
J.\ Rohmer, 
M.\ Schlueter, 
N.\ Urquhart, 
A.\ Zamuda 
and all anonymous contributors for their time and effort to fill in the questionnaire.

Timo M. Deist is funded by the 
 Open Technology Programme (project No. 15586), 
 financed by the Dutch Research Council (NWO), Elekta, and Xomnia. 
 This project is co-funded by the public-private partnership allowance for top consortia for knowledge and innovation (TKIs) from the Ministry of Economic Affairs.
 Boris Naujoks acknowledges the 
European Commission's H2020 programme, 
 H2020-MSCA-ITN-2016 UTOPIAE (grant agreement No. 722734) and the DAAD (German Academic Exchange Service), Project-ID: 57515062. 
Tea Tu\v{s}ar acknowledges financial support from the Slovenian Research Agency (projects No. Z2-8177 and BI-DE/20-21-019 and program No.\ P2-0209) and the European Commission's Horizon 2020 research and innovation program (grant agreement No. 692286).

\bibliographystyle{plain}
\bibliography{merge}

\backmatter
\appendix
\include{appendix}
\include{glossary}
\printindex


\end{document}